\documentclass[10pt,letterpaper,twocolumn]{article}

\usepackage{cvpr}
\usepackage{times}
\usepackage{epsfig}
\usepackage{graphicx}
\usepackage{amsmath}
\usepackage{amssymb}

\usepackage{algorithm}
\usepackage{algorithmicx}
\usepackage{algpseudocode}
\usepackage{epstopdf,enumitem}
\usepackage[pagebackref=true,breaklinks=true,letterpaper=true,colorlinks,bookmarks=false]{hyperref}
\cvprfinalcopy 

\DeclareMathOperator*{\argmax}{arg\,max}

\def\cvprPaperID{762} % *** Enter the CVPR Paper ID here
\ifcvprfinal\fi
\begin{document}

%%%%%%%%% TITLE
\title{Unsupervised Fusion Weight Learning in Multiple Classifier Systems}

\author{Anurag Kumar\\
Carnegie Mellon University\\
Pittsburgh-15213,USA\\
{\tt\small alnu@andrew.cmu.edu}
% For a paper whose authors are all at the same institution,
% omit the following lines up until the closing ``}''.
% Additional authors and addresses can be added with ``\and'',
% just like the second author.
% To save space, use either the email address or home page, not both
\and
Bhiksha Raj\\
Carnegie Mellon University\\
Pittsburgh-15213,USA\\
{\tt\small bhiksha@cs.cmu.edu}
}
\maketitle

%\thispagestyle{empty}

%%%%%%%%% ABSTRACT
\begin{abstract}
In this paper we present an unsupervised method to learn the weights with which the scores of multiple classifiers must be combined in classifier fusion settings. We also introduce a novel metric for ranking instances based on an index which depends upon the rank of weighted scores of test points among the weighted scores of training points. We show that the optimized index can be used for computing measures such as average precision. Unlike most classifier fusion methods where a single weight is learned to weigh all examples our method learns instance-specific weights. The problem is formulated as learning the weight which maximizes a clarity index; subsequently the index itself and the learned weights both are used separately to rank all the test points. Our method gives an unsupervised method of optimizing performance on actual test data, unlike the well known stacking-based methods where optimization is done over a labeled training set. Moreover, we show that our method is tolerant to noisy classifiers and can be used for selecting $N-$ best classifiers. 
\end{abstract}

\section{Introduction}
\subsection{Classifier Fusion}
In several pattern recognition tasks we are required to combine information from several sources. Fusion of the information derived from these sources thus becomes an important part of pattern recognition. Fusion may be performed {\em early}, by directly considering the features derived from the individual information sources jointly. {\em Late} fusion, on the other hand, is performed at the decision level, by somehow combining the decisions made from the information from the individual sources.  In this work our focus is on decision-level fusion, specifically of the variety where the final decision is based on a weighted sum of scores produced by individual classifiers. 
Optimization of fusion amounts to optimally learning the weights with which the classifiers are combined. Unlike the usual approach of learning a global set of weights that apply to all test instances, we attempt to learn  {\em instance-specific} weights for individual test points. Thus, we are able to consider the specific characteristics of individual data points, unlike global weighting schemes which completely ignore the individuality of test instances.

%Ensemble learning methods that combine several learning algorithms are known to outperform single-classifier learning. They have been analyzed theoretically and experimentally and shown to generally provide better results, compared to those obtained from a single classifier \cite{dietterich1}\cite{dietterich2}. Decision-level fusion methods can also be assumed to be a form of ensemble learning where different models are trained on different features or classification algorithms and finally combined to obtain the best results. They are desirable not only because individual classifiers may not give the required performance accuracy by themselves, and ensemble approaches are therefore required, but also because they are natural in several situations. Applications such as multimedia indexing and retrieval require fusion of information from different modalities such as text, video and audio; moreover, even within each of these modalities detection of objects or events requires extraction of several features, information from all of which must be combined. 
%In these applications, while early fusion strategies could be applied \cite{gehler}, they may be constrained by the fact that it might not be possible to combine the features directly at all ({\em e.g.} because of asynchronicity), and contrived methods of feature combination may actually have a deleterious effect on the learning. Consequently, late fusion strategies that combine scores predicted from several classifiers become important.

Most classifier-fusion methods assign a fixed weight to each classifier, the simplest and most common being the method of averaging which assigns equal weight to all classifiers. Even this simple averaging is quite effective and is often hard to beat in several situations, especially when different classifiers are almost independent.  Several other methods are discussed in the next sub-section. Nearly all of these methods rely on learning a unique set of weights from training or held-out data; these weights are subsequently used on all test instances, ignoring instance-specific behaviors of the classifiers. Moreover, questions relating to the generalization of weights learned from held-out data to the test data being scored also arise.
%However, in many situations such as medical diagnosis using images, accounting for instance-specific behaviors becomes important. The goal there is to determine the best strategy for a specific test instance. 
%Moreover, methods based on the ``stacking'' formalism \cite{wolpert} rely on labeled training or held-out data for learning weights. This also raises generalization concerns when the learned weights are applied to actual test instances \cite{kittler1}. Learning instance-specific weights on the actual test data can address these concerns. Hence, our objective in this work is to investigate and propose solutions for the instance-specific weight learning paradigm.  
%However, only one or two recent works \cite{ye} \cite{liu} have tried to look into instance specific weight learning. \cite{liu} in particular shows significant improvement in results in when instance specific weights are learned. 

In this work we propose solutions to learn instance-specific weights for classifier fusion, and investigate their behavior. We consider two scenarios.
\begin{itemize}[leftmargin=0.15in]
\item In the first, conventional, mechanism for combining classifiers, the final score of an instance is the weighted sum of {\em all} classifier scores. Thus all classifiers contribute to the final score of an instance.
\item In the second, only a subset ($N$-best) of classifiers contribute scores to an instance.  This lets us reject unreliable or noisy classifiers on a by-instance basis
\end{itemize}
Our method is based on the bipartite ranking loss \cite{freund}\cite{huang}. Two modifications of the bipartite ranking loss, called the relevance loss and the irrelevance loss, and an index defined on them are sufficient to learn optimal fusion weights for an unlabeled instance. Specifically, the idea is to optimize a raw ``clarity index'' with respect to the weights, to estimate the instance-specific fusion weights. Moreover, the optimal raw clarity indices of unlabeled instances can themselves be used to rank the unlabeled instances as well. 
The raw clarity index thus gives us an entirely novel mechanism of combining classifiers to score instances for ranking, which differs from the usual approach of ranking them by the weighted sum of the scores of the classifiers. 

Our method is unsupervised and the optimization of weights is done directly on actual test instances, rather than using a held out set. The method is unsupervised in the sense that in order to learn the optimal weights for an instance,  all we need are the scores from the classifiers for that instance, in addition to classifier outputs on training data. Since the optimization is performed directly on test data, it minimizes generalization concerns that result from learning weights on held out validation data \cite{kittler1}. As a corollary, the optimization is performed on actual test instances whose labels are not known and no intermediate held-out/validation data with known labels are required.

Ours is a {\em meta} algorithm; the training of the individual classifiers themselves is treated as a black box. We only consider the scores output by the classifiers. Our learning method does not know what kind of classifiers or features were used to obtain the scores. The only assumption we make is that within any classifier, higher scores imply a higher ranking of the instance by that classifier. From our perspective, an inverted ``bottom-up'' order of ranking -- where a {\em lower} score implies a higher rank -- is as good as a top-down ranking, provided the direction of ranking is known, since the former can be converted to the latter by a simple affine shift of the scores. Specifically, in our case the classifier scores are assumed to have the aspect of probabilities of belonging to the class; thus the higher it is for an instance the higher the rank of that instance among the set. 

In next subsection we give a brief description of some related work on classifier fusion. In Section 2 we describe the problem and our solution. In Section 3 we give experimental results using our method on object (flower) categorization. In Section 4 we discuss our results and conclude.

\subsection{Related Work}
Several works have studied the problem of classifier fusion \cite{ceamanos} \cite{junior}, \cite{singh03} \cite{lam} \cite{xu} \cite{ruta}. A particularly popular formalism for combining outputs of classifiers is stacking \cite{wolpert}. Stacking in general is implied in any method which involves ``learning'' to combine the base classifiers. The fundamental idea of stacking is that the problem of combining the base classifiers can be cast as another learning problem. The outputs (say probabilities) of the base classifiers are treated as an input space to the stacking function, while the output space of the function remains the same as that of the base classifiers \cite{mertz} \cite{todorovski}. The stacking framework learns the parameters of the stacking function to optimize classification accuracy, generally on some labeled training or held-out data.  Our approach, on the other hand,  does not optimize classification accuracy -- the objective that is optimized is an index called \emph{clarity}, and makes {\em no} reference to the true labels of the data that it is optimized over. The combination function is optimized in an unsupervised manner over the actual test data. Moreover, we preform the optimization separately for each test instance.

To the best of our knowledge, few recent works have actually looked into instance-specific weight learning \cite{liu} \cite{ye}. Some of the most promising results are reported in \cite{liu}. The basic idea in this work is to propagate fusion weights of labeled instances to the individual unlabeled instances along a graph built on low-level features. The method has been shown to outperform other fusion methods on a variety of datasets. However, although the learned weights are instance specific, the method not only still requires a held-out set for which labels are known, it also requires knowledge of the low-level features of instances. On the other hand, our method does not require held-out data. Moreover, our solution is a {\em meta} algorithm that requires no knowledge of the low-level features of the instances.
Another issue with \cite{liu} is that the weights learned for different test instances are not disjoint from each other. This has the undesirable aspect that newer test instances cannot be independently introduced into the set. 
%Thus, while the work does demonstrate that instance-specific optimization of fusion weights can outperform other methods, other desirable aspects of instance-specific weight optimization, such as the ability to process test instances independently, or learning to select the $N$-best classifiers for an instance are missing. 

Given the distinctness of our approach, we focus  on introducing and investigating our proposed instance-specific weight-learning paradigm, rather than demonstrating improvements over several other global fusion strategies. Unlike the other methods mentioned earlier, our solution does not require a separate held-out set. Also,  the optimization of weights for each test instance is disjoint from other test instances. Finally, our method is as true {\em meta} algorithm that makes no reference to low-level features or how the classifiers were trained. 

We also analyze important aspects of the fusion such as selecting only a group of good classifiers for an instance and the effects of noisy classifiers on the weight learning scheme and show that our proposed method is quite robust.
\section{The Proposed Algorithm}
\label{gen_inst}

\subsection{Problem Setting}
We set up our problem within a retrieval scenario where the objective is to rank positive test instances from the target class ahead of negative instances. Our objective becomes that of determining how  to combine the scores produced by a collection of classifiers, in order to optimize the ranking.

Let $p$ be a sample instance and $m$ be the number of classifiers used for predicting scores. $C_i$ denotes the $i^{th}$ classifier. Thus for any sample instance $p$ we have $m$ outputs scores $x_i=C_i\left(p\right),\, i=1 \cdots m$, where $C_i\left(p\right)$ is the output of classifier $C_i$ on some feature vector of $p$. Let $\vec{x} = [x_1\,x_2\, x_3\,.....\,x_m]^T$ be the vector representing the scores from all $m$ classifiers. Thus all sample instances are represented by an $m$-dimensional score vector. Let $y \in \{0,1\}$ represent the label of an instance.

Let $X$ be the set of available training instances, where each instance in $X$ is represented by an $m$-dimensional score vector. Class labels $y$ are available for every instance in this set. We note that $X$ here represents the set that will be used to optimize the fusion, not the data used to train the $m$ individual classifiers. In practice, it is sufficient to have $X$ be the same as the training set on which all classifiers are trained, and hence no held-out set is required in the learning process. However, mathematically no such restriction is placed. The positive ($y=1$) and negative ($y=0$) labeled training instances in $X$ are separated into two sets, $X_+$ and $X_-$, such that each instance in $X_+$ has label $y=1$ and each instance in $X_-$ has label $y=0$. The number of instances in $X_+$ is represented by $n_1$ and in $X_-$ by $n_0$. 

Let $X_{test}$ be the set of unlabeled test instances that must be classified and $p^u$ be an unlabeled test instance in $X_{test}$ with score vector $\vec{x}^u$. The goal is to learn an optimal weight vector $\vec{w}_u$ for each unlabeled instance $p^u$ and the final weighted sum of scores for each $p^u$ given by $s^u=\vec{w}_u^T\vec{x}^u$. 

\subsection{Relevance, Irrelevance and Clarity}
\begin{figure}[h!]
  \centering
    \includegraphics[width=0.4\textwidth]{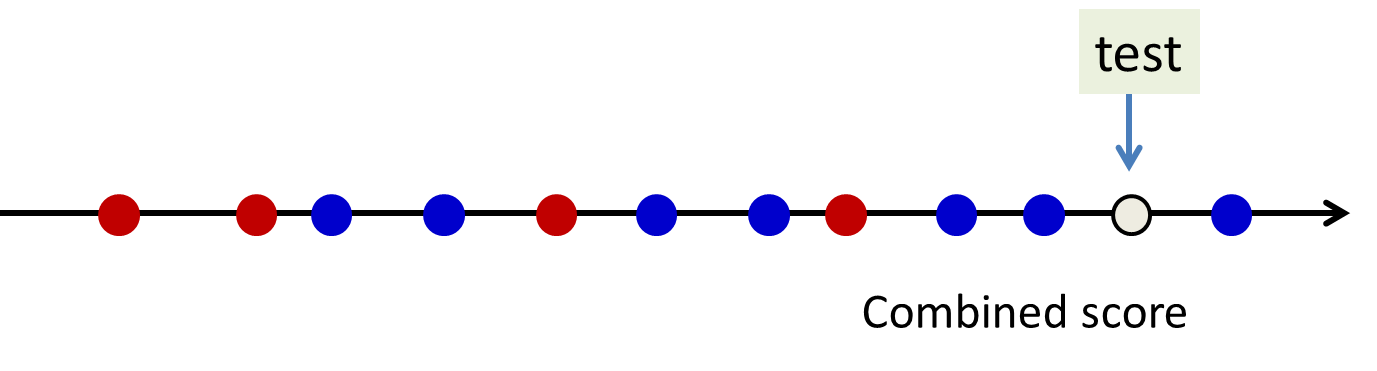}
		  \caption{The axis represents combined score of instances. The red dots represent negatively labeled training instances. The blue dots are positively labeled training instances. The test instance is shown by the grey dot. 6 of 7 positive instances score less than the test instance, hence the irrelevance loss is $6/7$.  None of the negative instances score more than the test instance, hence the relevance loss is 0. The clarity is $|0-6/7| = 6/7$.}
			\label{fig:clarity}
\end{figure}
The fusion weights are learned directly on test instances for which the class labels are not known. To learn the weights, we must define an objective function that does not refer to the labels. Instead, our objective will relate to the rank of the test instance relative to the training instances.

For each test instance $p_u$ we aim to find the weight vector $\vec{w}_u$ that maximizes the score $s^u$ if the instance is positive, or minimizes (makes it maximally negative) it if its negative. In order to do so, we  define an objective function that, when optimized, can be expected to result in weights that have these characteristics.  We do so as follows.\footnote{Note that we assume here, without loss of generality, that the classification rule assumes the score to be analogous to the probability of belonging to the target class -- higher scores imply a higher probability and vice versa.}

We base our objective on the intuition that if $p_u$ is to be classified well, then, if the test instance is positive, its score must lie as far to the right of the distribution of the scores of positively labeled instances as possible for confident classification. Empirically, the instance must outscore as many of the positively labeled training points in $X_+$ as possible. On the other hand, if the instance is negative, its score must ideally be lower than that of as many negatively labeled training instances from $X_-$ as possible.

To formalize this intuition, we define two losses, the \emph{relevance loss} and the \emph{irrelevance loss}, and an index based on these losses \cite{huang}. The \emph{relevance loss} $RL(\vec{x}^u,\vec{w}_u)$ for an unlabeled point $p^u$ with score vector $\vec{x}^u$ and weight vector $\vec{w}_u$ in our setting is defined as the fraction of negatively labeled training instances from $X_-$ that score {\em more} than $p_u$, when the scores are combined using $\vec{w}_u$:
\begin{equation}
RL(\vec{x}^u,\vec{w}_u)=\frac{1}{n_0} \sum\limits_{i=1}^{n_0} I\left(\vec{w}_u^T\vec{x_i}-\vec{w}_u^T\vec{x}^u\right) \,\,\, \forall\,\, \vec{x_i} \,\in\,X_-
\end{equation} 
where $I$ is the indicator function such that 
\begin{center}
$I(t) =
\begin{cases}
1, & \text{if }t \geq 0 \\
0, & \text{otherwise }
\end{cases}$
\end{center}

Similarly, the \emph{irrelevance loss} $IL(\vec{x}^u,\vec{w}_u)$  is defined as the fraction of positively labeled training instances from $X_+$ that score {\em less} than $p_u$%, when $\vec{w}_u$ is used as the weight vector.
\begin{equation}
IL(\vec{x}^u,\vec{w}_u)=\frac{1}{n_1} \sum\limits_{i=1}^{n_1} I\left(\vec{w}_u^T\vec{x}^u-\vec{w}_u^T\vec{x_i}\right)\, \forall \, \vec{x_i} \in \,X_+
\end{equation}

If the unlabeled instance $p^u$ has a true label $y_{p^u}=1$, it is desired that its \emph{relevance loss} be low (0 in the ideal case). Also, the higher the \emph{irrelevance loss} the more confidence we have for $p^u$ to be positive. However, if $p^u$ is actually a negative point, {\em i.e.} $y_{p^u}=0$ then the \emph{irrelevance loss} should be very low, whereas the higher the value of the \emph{relevance loss}, the higher is our confidence on $p^u$ being a negative instance. These two factors can be combined into a single index termed as \emph{Clarity Index}. The \emph{clarity index} is defined as the absolute value of the difference between the \emph{relevance loss} and \emph{irrelevance loss}.
\begin{equation}
CL(\vec{x}^u,\vec{w}_u)=|RL(\vec{x}^u,\vec{w}_u)-IL(\vec{x}^u,\vec{w}_u)|
\end{equation}
Figure \ref{fig:clarity} illustrates the relevance and irrelevance losses and the clarity index.
It is obvious that the higher the value of  the \emph{clarity index}, the easier it is to make a decision for  $p^u$. The range of the clarity index is $[0,1]$ and it is desired for it to be high for any unlabeled instance. 
%This facilitates in using this index as a measure to decide the query in active learning setting\cite{huang}.
We also define the \emph{Raw Clarity Index (RCL)} which is just the difference between \emph{RL} and \emph{IL}. Thus $RCL(\vec{x}^u,\vec{w}_u)=RL(\vec{x}^u,\vec{w}_u)-IL(\vec{x}^u,\vec{w}_u)$ and the range of $RCL$ is $[-1,1]$. $CL$ is the absolute value of $RCL$. For a positive instance we expect the \emph{raw clarity index} to be negative; the closer it is to $-1$ the better it is. Similarly for a negative instance the desired value $RCL$ is to be positive and high. In all cases, the $CL$ value should be high. This \emph{raw clarity index}, as we describe in a subsequent subsection, can also be used as another way to rank the test instances along with the weighted sum of scores $s_u$.

However, direct optimization of $CL$ with respect to $\vec{w}_u$ is intractable in general, because the function $I$ in the definitions of $RL$ and $IL$ is a discrete measure and cannot be differentiated. We approximate it instead by a smooth, differentiable sigmoid function: 
\begin{equation}
I(t) \approx I_s(t)=\frac{1}{1+e^{-\alpha t}}
\label{eq:sigmoid}
\end{equation}
By choosing the correct $\alpha$ this function can be made arbitrarily close to  the indicator function $I$.

Using this approximation, the \emph{relevance loss (RL)} and \emph{irrelevance loss}(IL) are redefined as
\begin{eqnarray}
RL(\vec{x}^u,\vec{w}_u)=\frac{1}{n_0} \sum\limits_{i=1}^{n_0} \frac{1}{1+e^{-\alpha \vec{w}_u^T \left(\vec{x_i}-\vec{x}^u\right)}} \,\,\, \forall\,\, \vec{x_i} \in\,X_- \\
IL(\vec{x}^u,\vec{w}_u)=\frac{1}{n_1} \sum\limits_{i=1}^{n_1} \frac{1}{1+e^{-\alpha \vec{w}_u^T \left(\vec{x}^u-\vec{x_i}\right)}} \,\,\, \forall\,\, \vec{x_i} \in\,X_+
\end{eqnarray}

%The $\alpha$ which gives the best approximation will depend on the dataset. This is shown for a single point in two dimensional setting with the restriction of $w^Tw=1$ in \ref{fig:1}. The restriction allows us to vary just the angle between $0$ to $\pi/2 $ and plot actual(using function $I(t)$) and the estimated clarity indices.  (a) correspond to synthetic data where scores for $X$ and $\vec{x}^u$ are randomly generated. (b) corresponds to real data where data in $X$ and $\vec{x}^u$ is from the database used for experiments. 

%\begin{figure}
% \centering
%   \includegraphics[width=\columnwidth]{syhthe.png}
%  \caption{Clarity Index vs. Angle for synthetic (a) and real(b) data}
%  \label{fig:1}
%\end{figure}

\subsection{Learning Weights}
We now present a method to learn the instance-specific fusion weights.  Our goal is to finding the weight that maximizes the clarity index. The clarity index CL is the absolute value of the raw clarity index RCL. The absolute value function, like the indicator function, is non-differentiable at $0$. We may bypass this by employing a continuous, differentiable, approximation of the absolute value function; however, we employ the following direct strategy instead.

The \emph{raw clarity index} using the sigmoid functions is 
\begin{equation}
\label{rclfun}
\begin{split}
RCL(\vec{x}^u,\vec{w}_u)=\frac{1}{n_0} \sum\limits_{\forall \vec{x_i} \in X_-} \frac{1}{1+e^{-\alpha \vec{w}_u^T \left(\vec{x_i}-\vec{x}^u\right)}} \\
-\frac{1}{n_1} \sum\limits_{\forall \vec{x_i \in X_+}} \frac{1}{1+e^{-\alpha \vec{w}_u^T \left(\vec{x}^u-\vec{x_i}\right)}}
\end{split}
\end{equation}
Since $CL = |RCL|$ we can maximize CL as:
\begin{eqnarray}
\label{optim}
\vec{w}_{max} &=& \arg\max_{\vec{w}_u} RCL(\vec{x}^u,\vec{w}_u) \nonumber \\
RCL_{max} &=& RCL(\vec{x}^u, \vec{w}_{max}) \nonumber\\
\vec{w}_{min} &=& \arg\min_{\vec{w}_u} RCL(\vec{x}^u,\vec{w}_u) \nonumber \\
RCL_{min} &=& RCL(\vec{x}^u, \vec{w}_{min}) \nonumber
\end{eqnarray}
\begin{equation}
\hat\vec{w}_u =
\begin{cases}
\vec{w}_{max},~~if~~RCL_{max} > |RCL_{min}| \\
\vec{w}_{min}~~~otherwise
\end{cases}
\end{equation}
In other words, we estimate both the maximum and minimum values of $RCL$, and choose the weights corresponding to whichever of the two has the larger absolute value.

The above estimate requires both maximization and minimization of $RCL(\vec{x}_u, \vec{w}_u)$. We find these extrema through a gradient descent/ascent procedure.  Starting with some initial weight we estimate the maximum of $RCL$ with respect to $\vec{w}$ by  gradient ascent. We employ gradient {\em descent} from the same initial location to find the weight which minimizes $RCL$. Additionally, the weights are subject to constraints of $\vec{w} > 0$, since we assume all classifiers to be no worse than random. In addition, to keep the weights from exploding we also impose constraints of $||\vec{w}||^2=\vec{w}^T\vec{w}=1$, giving us a feasible set that lies on the surface of the section of a unit hypersphere that lies in the positive orthant. The weights are projected on the feasible region after each gradient descent/ascent step. Note that in general $RCL$ is not convex and the algorithms may get stuck in local optima in either direction.

The overall  algorithm for learning the weight for an instance $p^u$ is given in Algorithm 1. In Algorithm 1 $\frac{dRCL(\vec{x}^u,\vec{w}_{max})}{d\vec{w}_{max}}$ represents the derivative of $RCL$ defined in Equation \ref{rclfun} w.r.t $\vec{w}_{max}$. $\eta_k$ is the ascent step size for the $k^{th}$ iteration which can be fixed or chosen by any search method for each iteration. Similar definitions apply for the minimization case.

Computationally, convergence to local optima is pretty fast. In most test cases in our experiments the algorithm quickly converges and no significant load is observed in spite  of the method being an instance specific approach. 
%One might be concerned about scalability of the method to large scale test cases. However, in these situations methods which learn single set of weights using the training or a small held-out  set will not be able to keep with the large variations in scores of classifiers on In comparison with other methods where optimiza 

\begin{algorithm}
\caption{Weight Learning Algorithm}\label{euclid}
\begin{algorithmic}[1]
\Procedure{Learning Weight for each $p^u$}{$X_+,X_-,,\vec{x}^u$} // Input training score vectors and score vector of $p^u$
\newline\newline
//Obtain weight which maximizes $RCL$
\State Initialize $\vec{w}_{max}\gets [w_1\,w_2\,w_3\,....w_m]^T$, $w_i \ge 0~ \forall i$ 
\Repeat
%\While{}
\State $\vec{w}_{max} \gets \vec{w}_{max}+\eta_{k} \frac{dRCL(\vec{x}^u,\vec{w}_{max})}{d\vec{w}_{max}}$ %\Comment{$\frac{dRCL(\vec{x}^u,\vec{w}_{max})}{d\vec{w}_{max}}$ represents derivative of $RCL$ defined in Equation \ref{rclfun} w.r.t $\vec{w}_{max}$ and $\eta_k$ is the step for $k^th$ iteration}
\State {\small Project onto $\{\vec{w}~:~\vec{w}_{max} > 0~\&~||\vec{w}_{max}||^2=1\}$}
\Until Convergence of $RCL$
\newline\newline
//Now Obtain Weight which minimizes $RCL$
\State Initialize $\vec{w}_{min}\gets [w_1\,w_2\,w_3\,....w_m]^T$, $w_i \ge 0~ \forall i$
\Repeat
%\While{}
\State $\vec{w}_{min} \gets \vec{w}_{min}-\eta_{k} \frac{dRCL(\vec{x}^u,\vec{w}_{min})}{d\vec{w}_{min}}$ %\Comment{$\frac{dRCL(\vec{x}^u,\vec{w}_{min})}{d\vec{w}_{min}}$ represents derivative of $RCL$ defined in Equation \ref{rclfun} w.r.t $\vec{w}_{min}$and $\eta_k$ is the step for $k^th$ iteration}
\State {\small Project onto $\{\vec{w}~:~\vec{w}_{max} > 0~\&~||\vec{w}_{max}||^2=1\}$}
\Until Convergence of $RCL$
\newline\newline
//Assign $\vec{w}_u$ to weight for which absolute value of raw clarity or clarity index is higher 
%\small
\State {\small $\vec{w}_u = \argmax_{\vec{w}}\,(|RCL(\vec{x}^u,\vec{w}_{min})|,|RCL(\vec{x}^u,\vec{w}_{max})|)$}
%\normalsize
\State Clarity($p^u$) = $RCL(\vec{x}^u,\vec{w}_{u})$
%\EndWhile\label{euclidendwhile}
%\State \textbf{return} $b$\Comment{The gcd is b}
\EndProcedure
\end{algorithmic}
\end{algorithm}

\subsection{Ranking Instances}

%==================================

% The core idea of our method is to find the weight vector which maximizes the clarity index ($CL$) for the instance $p^u$. This weight vector $\vec{w}_u$ is the best weighing through which we will have the maximum confidence on the decision made about $p^u$. This is done for each unlabeled instance separately; thus the behavior of the classifiers on a unique $p^u$ is captured, and the weight vector for one unlabeled instance is not distorted because of some other unlabeled instance on which some of the classifiers might have performed badly. Thus by point-specific weight learning we not only consider the uniqueness of that point but also removes any aberrations in weight learning due to predictions on other instances, potentially resulting in a more robust classification.

%Moreover, since for every unlabeled instance the weights are updated with respect to labeled instances in $X_+$ and $X_-$ even if the individual classifiers  had done a bad job on this instance we will still have the weighing vector on which we have maximum confidence. 
%Finally, all unlabeled instances are ranked by the weighted sum of scores where the weighting is performed by the weight vector learned for each individual instance. This is feasible because the optimization of weights of each instance is performed using the same set $X$.

The algorithm of Algorithm 1 results in the estimation of fusion weights for each test instance. These can now be used to compute scores and rank order the set of test instances in a number of ways.

\noindent{\bf Ranking by weighted score}  The estimated weights can simply be used to compute the score $s_u$ for every test instance according to the weighted score $s_u = \vec{w}_u^T \vec{x}_u$.

\noindent{\bf Ranking by raw clarity index:} 
We also introduce another ranking method based on the \emph{raw clarity index}. Since we optimize the raw clarity index, the raw clarity itself is a measure for ranking of unlabeled instances. As discussed previously,  for a positive instance we expect the \emph{raw clarity index} to be negative; the closer it is to $-1$ the better, and for a negative instance the desired value $RCL$ is to be close to $+1$. After optimization whatever value $CL(\vec{x}^u,\vec{w}_u)$ stores is the best that can be achieved in either of the two directions. Hence we can simply rank the unlabeled instances based on the reverse order of their optimal \emph{raw clarity index}. In our experiments we show that this ranking method can sometimes actually result in better ranking compared to that based on the weighted score $s_u$ values. Thus we also now have a novel metric for ranking instances, which is based on the optimal {\em rank} of the test instances, rather than their weighted-combined score. 

\subsection{$N$-best Selection}
We note that poor or noisy classifiers can have a detrimental effect to the overall classification. Classifiers are usually trained on a limited amount of training data. Test instances of unseen characteristics can hence evoke erratic behavior from a classifier. Since the classifiers have been trained on the training data, the probability of such behavior will be low on the training data itself. As a result, the score assigned to a test instance may not be well explained by the distribution of scores obtained on the training data. The weight-learning algorithm should ideally be able to identify such detrimental classifiers and assign very low weight to them. In effect, the estimated weights effectively assign an {\em importance} to each of the fused classifiers; noisy or mismatched classifiers should obtain low weight in the weight optimization process. 

We can therefore use the proposed method to select the $N$-best classifiers to judge any test instance, by selecting the classifiers corresponding to the highest $N$ weights. 

In this $N$-best scenario, ranking can subsequently be done in one of several ways: 
\begin{enumerate}
\item By simple averaging of the $N$-best classifiers.
\item By computing the weighted score $s^u_N$ over the $N$-best classifiers, employing the already estimated weights.
%\item By computing an $N$-best raw clarity for the instance based on the $N$-best weighted score, $^u_N$.
\end{enumerate}

\section{Experimental Results}
\label{headings}
We evaluate the performance of our method on multiclass object categorization. We use the Oxford Flower dataset \cite{nilsback1} which has been used in several works such as \cite{gehler}\cite{nilsback1}\cite{nilsback2} to name a few. This dataset contains flowers of 17 different categories. It provides 80 images for each flower class resulting in an overall set of 1360. The dataset has three predefined splits. In each predefined split, all flower classes are split into $40$ training images, $20$ validation images and $20$ test images. The dataset also provides $7$ different features for the images. \cite{nilsback1} describes the details of features based on Colour Vocabulary, Shape Vocabulary and Texture Vocabulary. \cite{nilsback2} gives the details of features based on HSV, SIFT on the foreground internal regions, SIFT on the foreground boundary, and Histogram of Gradients. The $\chi^2$ distance matrix for all 7 features are also provided. The predefined splits are here referred to as \emph{SET1, SET2, SET3}. 

Our basic classifiers are $\chi^2$ kernel based SVM classifiers. For each flower class we train $7$ different base SVM classifiers corresponding to the $7$ different features in one-versus-rest fashion. Experiments are done as per the predefined splits. The best parameters for the SVM classifiers are chosen by performance check on the validation set. The outputs of these base classifiers on the specified training set forms the training set $X$ for our fusion method and the outputs corresponding to specified test set form our test set $X_{test}$. Each instance is thus represented by a $7$-dimensional score vector corresponding to the outputs from $7$ different classifiers. 

Since our focus here is more on analysing the unsupervised instance-specific learning paradigm, which presents a new take on fusion strategies, and for which no truly-equivalent comparator exists, we compare our performance with {\em average} fusion (AVG.). This is one of the most commonly used fusion schemes, where the final score is just the average score of all classifiers and can be very hard to beat specially when the performances of individual classifiers are high, as is true in the current case. We consider various aspects of the problem, including basic classification,  $N$-best selection and the performance of our method when noise is deliberately added to the classifiers. 

We report results in terms of {\em average precision} (AP) and {\em mean average precision} (MAP), which are effective characterizations of the accuracy of ranked lists, since, from our perspective, this is a retrieval task.  For any class, the AP for a list is given by
\[
AP = \frac{\sum_{i=i}^n P(i)I_+(i)}{N_+}
\]
where $N_+$ is the number of positive instances in the test set, $I_+(i)$ is an indicator of whether the $i^{\rm th}$ test instance is a positive instance for the class, and $P(i)$ is the fraction of the top-ranked $i$ instances which are positive. The MAP is the average of the AP of all classes in the test.

In all experiments we fixed the learning rate $\eta$ in Algorithm 1 as 0.1. 

\subsection{Selecting $\alpha$}
The sigmoid approximation of the indicator function given in Equation \ref{eq:sigmoid} has a key parameter $\alpha$. Setting this to a high value results in closer approximation to the true indicator function, but results in several local optima of the objective function ($CL$), effectively increasing the variance of the estimator.  A low value of $\alpha$, on the other hand, results in lower variance, but can have significant bias. Consequently, the actual value chosen for $\alpha$ can have a considerable effect on the outcome of the classifier.
\begin{figure}[h!]
\centering
\includegraphics[width=0.3\textwidth, height=1.5in]{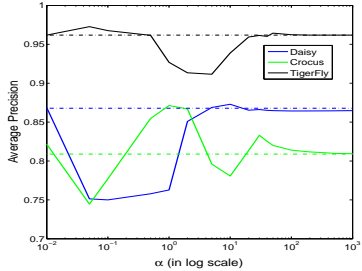}
\caption{AP as a function of $\alpha$ for three classes. The horizontal dotted lines show the AP with average fusion.}
\label{fig:alpha}
\end{figure}
Figure \ref{fig:alpha} shows the variation in AP as a function of $\alpha$ for three flower classes. As can be seen, there can be considerable variation in performance with $\alpha$.  In all cases, the performance obtained with the best $\alpha$ is significantly higher than that obtained with average fusion. 
%This can also be seen in Figure \ref{fig:mapposs} where the overall map is shown for $\alpha$ oracle provided alpha. This means true $\alpha$ for every test class is assumed to be known. 

For subsequent experiments, we set the $\alpha$ used for any class by optimizing performance on the specified validation sets in the data.

\subsection{Ranking by total score}
\begin{figure}[t]
\centering
\includegraphics[width=0.3\textwidth,height=1.5in]{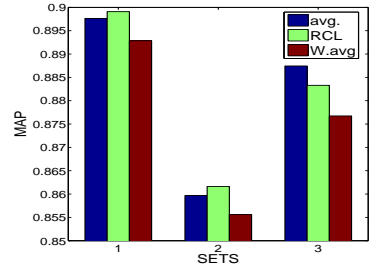}
\caption{MAP results on all three sets, for average fusion (avg.), weighted average using clarity-optimized weights (w.avg.), and optimized raw clarity (RCL).}
\label{fig:allclasses}
\vspace{-0.2in}
\end{figure}

We now report the performance obtained from the combined scores, where all 7 classifiers were combined. Figure \ref{fig:allclasses} shows the MAP performance over all 17 classes on the data set. Figure \ref{fig:allclasses} shows results on all three sets of the data.
The figure shows results obtained using three methods:  ranking with scores obtained from average fusion,  with scores  from weighted fusion using the optimized weights, and based on the optimized raw clarity.

\begin{figure}[t]
\centering
\includegraphics[width=0.3\textwidth,height=1.5in]{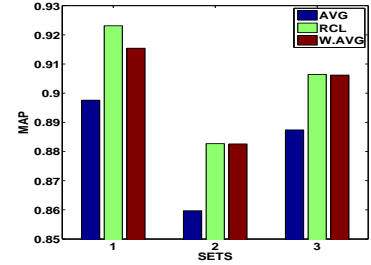}
\caption{MAP results on all three sets, for average fusion (avg.), weighted average using clarity-optimized weights (w.avg.), and optimized raw clarity (RCL) with oracle $\alpha$}
\label{fig:mapposs}
\vspace{-0.2in}
\end{figure}

Its interesting to note that ranking based on raw clarity outperforms weighted fusion in every case; in fact the latter is poorer than average fusion in this test (we see in the next section that this is not always so). Raw clarity based scoring also outperforms average fusion in two of the three sets. To reiterate the effect of $\alpha$ on the performance, we show the MAP values when $\alpha$ has been provided by an oracle in Figure \ref{fig:mapposs}. This essentially means $\alpha$ is tuned on test data. From Figure \ref{fig:mapposs} and Figure \ref{fig:alpha} we make a note of the fact that proper $\alpha$ tuning can give significant improvements. Apart from Figure \ref{fig:mapposs} all results are on $\alpha$ selected using specified validation sets.  
\subsection{$N$-best selection}
Not all classifiers that are fused are equally effective on any instance.  As mentioned earlier, the proposed weight-estimation strategy can actually be used to {\em select} the best classifiers for each instance.

\begin{figure}[t]
\centering
\includegraphics[width=0.225\textwidth,height=1.5in]{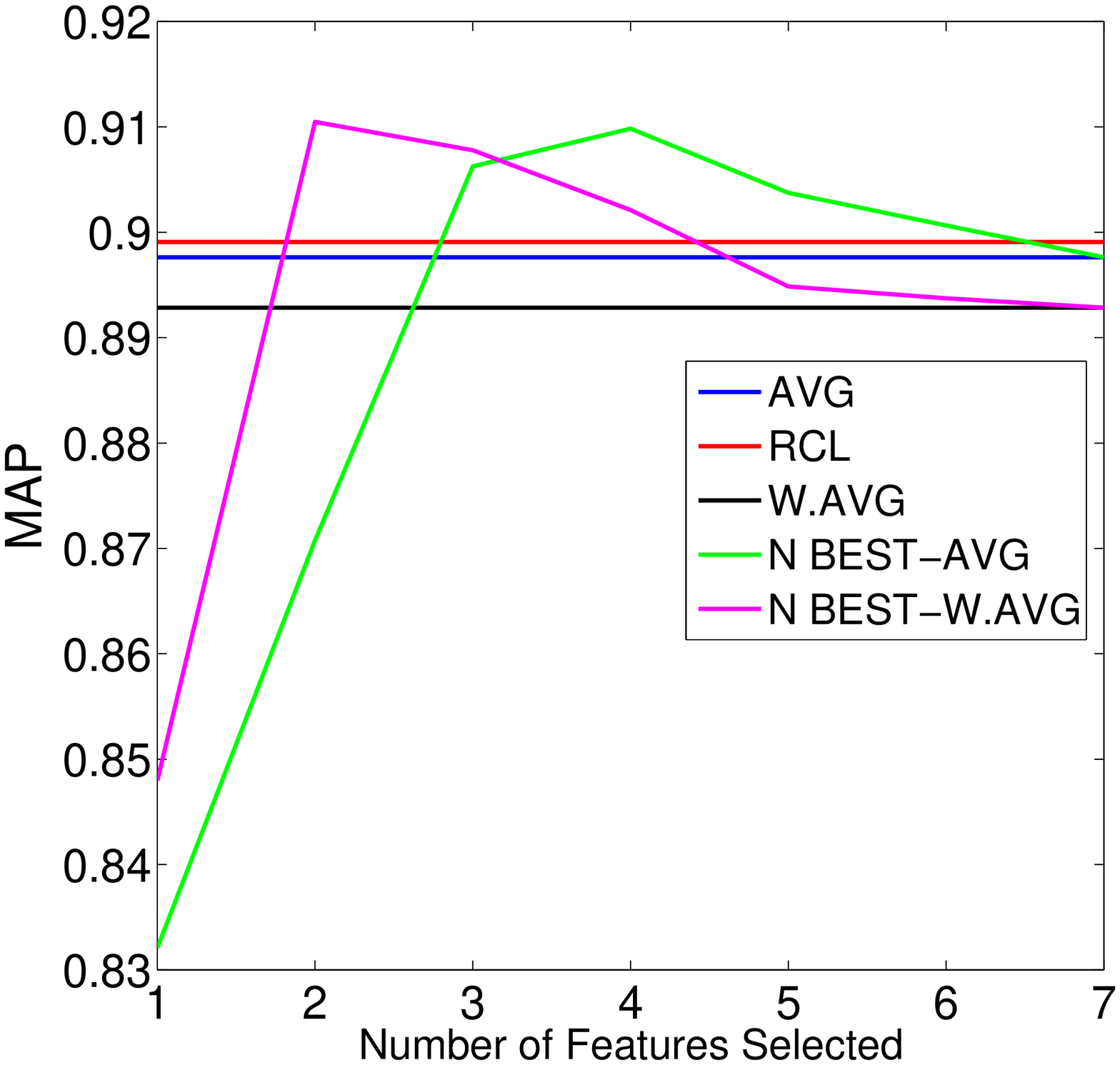}
\includegraphics[width=0.225\textwidth,height=1.5in]{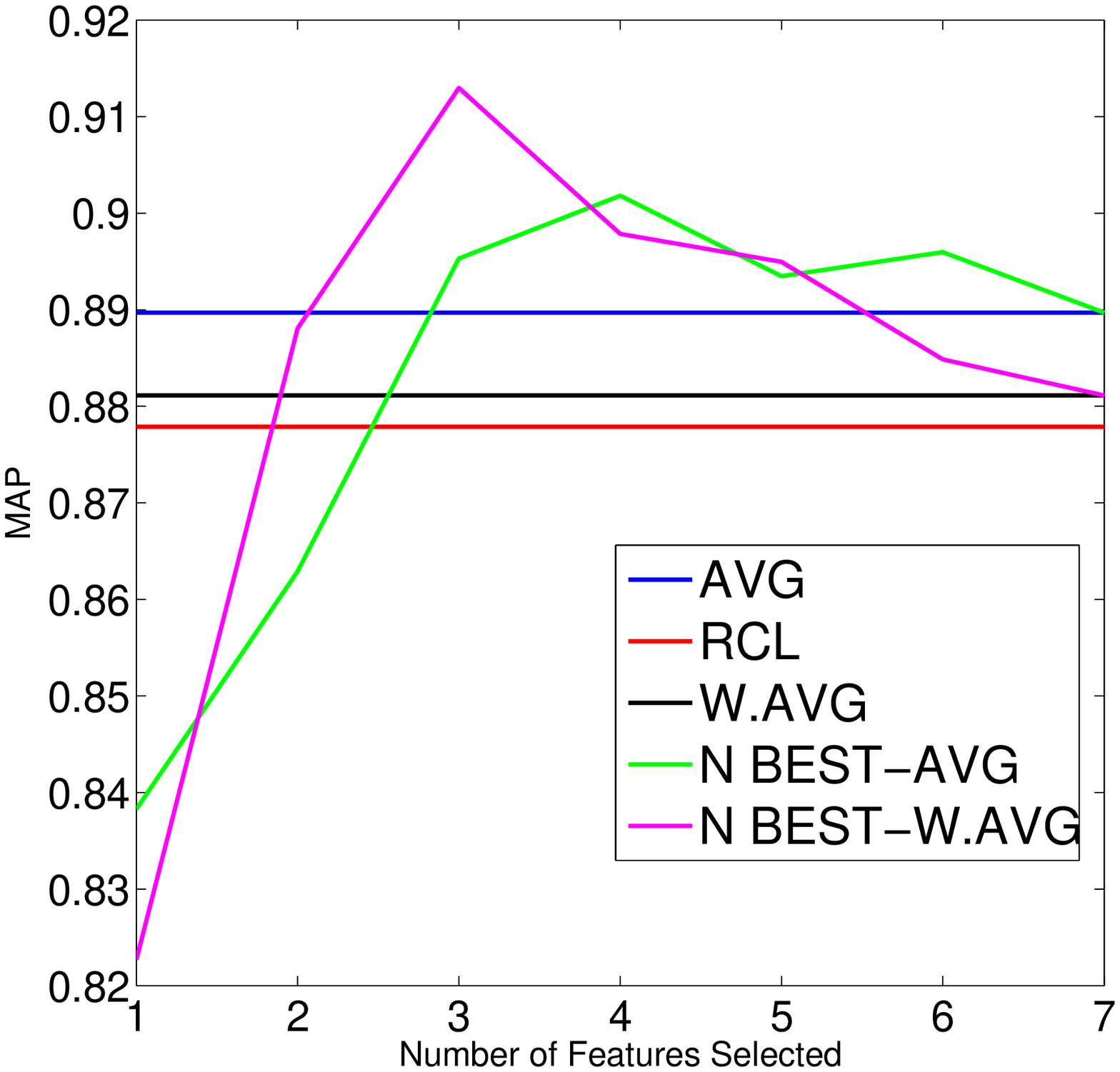}
\caption{MAP results on Set 1 with average fusion of $N$-best scores ($N$-BEST AVG) and weighted fusion of $N$-best scores ($N$-BEST WAVG), as a function of $N$. (a) Left: on clean data. (b) Right: when one of the classifiers is corrupted by Gaussian noise. }
%Horizontal lines show average fusion (avg.), clarity-optimized weighted average  (w.avg.), and optimized raw clarity (RCL).}
\label{fig:nbest}
\vspace{-0.18in}
\end{figure}

The left panel in Figure \ref{fig:nbest} shows the result of this approach on Set1. Results for the remaining two sets are submitted as part of the supplemental material. The figure shows the performance obtained with two variants, (1) the top $N$ classifiers are uniformly averaged, and (2) the weighted summed scores of the top $N$ classifiers is considered for ranking. The figure shows the performance as a function of $N$. The horizontal lines in the figure also show the performance obtained when all 7 classifiers are combined. We note that rejecting the worst scoring classifier improves the performance of both $N$-best approaches. Further, weight-based selection of classifiers can result in significant improvement over combining all seven classifiers. Here, superior performance is obtained in both, averaged $N$-best scores and weighted average of $N$-best scores. It may be noted from the supplemental material that even on the remaining sets, including the difficult Set 3, the performance with averaged $N$-best scores can be superior to all other methods for the appropriate setting of $N$.
\subsection{$N$-best selection on Noisy Classifiers}
To study this phenomenon of $N$ best selection in a clearer way we look into a harder problem. We deliberately introduce noise in the classifiers and observe if our weight learning algorithm can sustain this corruption by noise. So a classifier is artificially degraded by the addition of Gaussian noise to the scores given by the classifier on the test points. This is done only for a  percentage of randomly chosen test points. This simulates the effect of an erroneous classifier that may have been added to the mix. Such a classifier can badly affect the performance of any fusion scheme that combines all classifiers. This process can be done for more than one classifier as well. These noisy scores are then used to learn weights and we assess how much performance in terms of MAP has been sustained in this noisy situation. For observable decay of performance, in the present experiments $20\%$ of test points are corrupted for a classifier. The number of classifiers corrupted is denoted by $c$.  We perform experiments by corrupting $c=1$, , $c=3$ and $c=4$ classifiers. The classifiers to corrupt are chosen randomly. Having more noisy classifiers means more degradation in performance in terms of MAP. Our goal is to see if we can sustain the performance by the $N$-best selection schema which uses the weights learned by the proposed algorithm to select the $N$ best classifiers. In the right panel in Figure \ref{fig:nbest}, one of the classifiers has been artificially degraded. We note that $N$-best based methods remain robust to the inclusion of such degraded classifiers in the mix. This difference becomes more visible when number of such noisy classifiers is increased. \footnote{Note here that in this noise tolerance study, for $\alpha$ selection using validation set, even the validation set was corrupted. This was done to ensure that validation set is considered no different from test set and hence $\alpha$ must be selected based on corrupted validation data. This in fact proves robustness of our method.}

\begin{figure}[t]
\centering
\includegraphics[width=0.225\textwidth,height=1.5in]{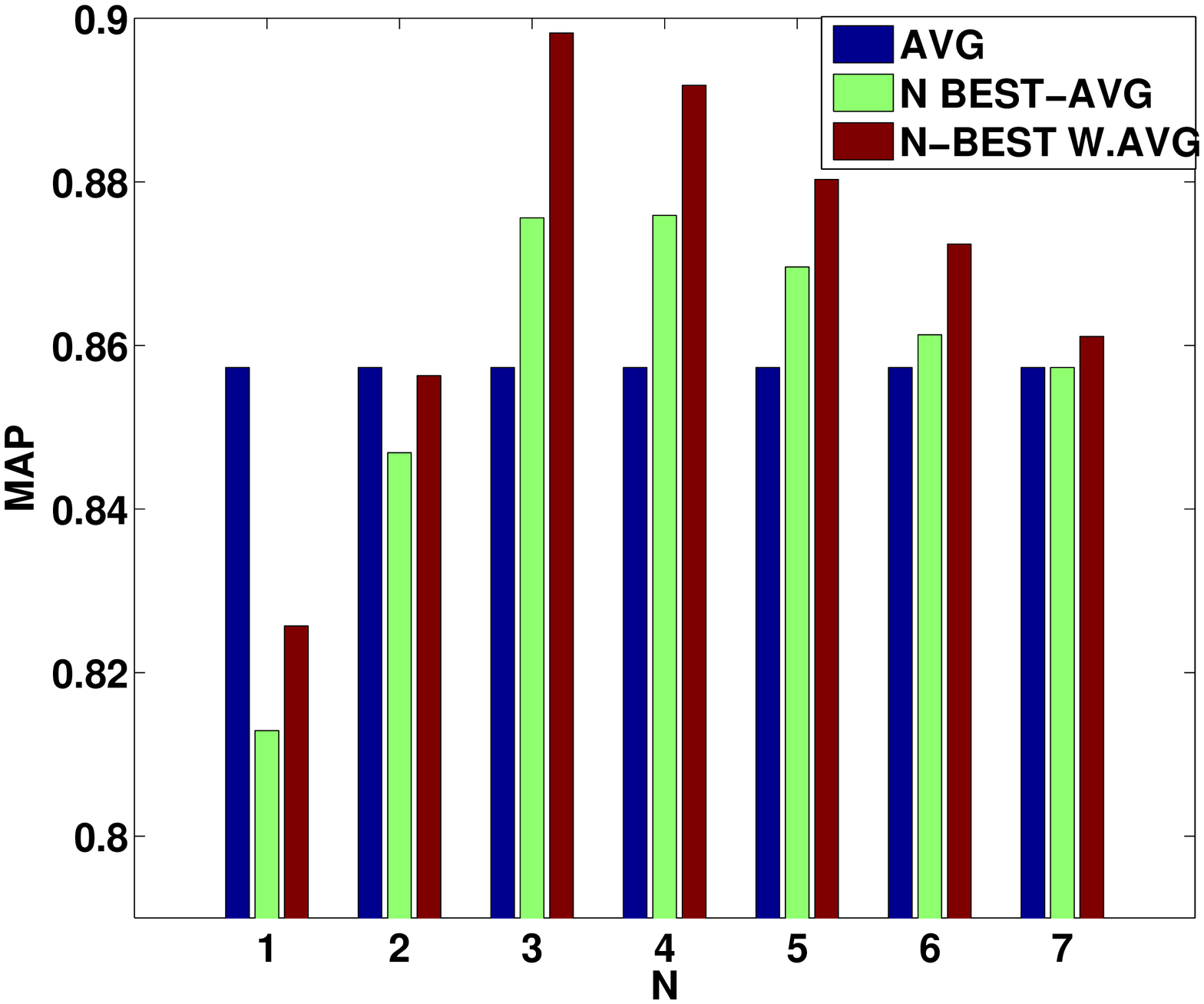}
\includegraphics[width=0.225\textwidth,height=1.5in]{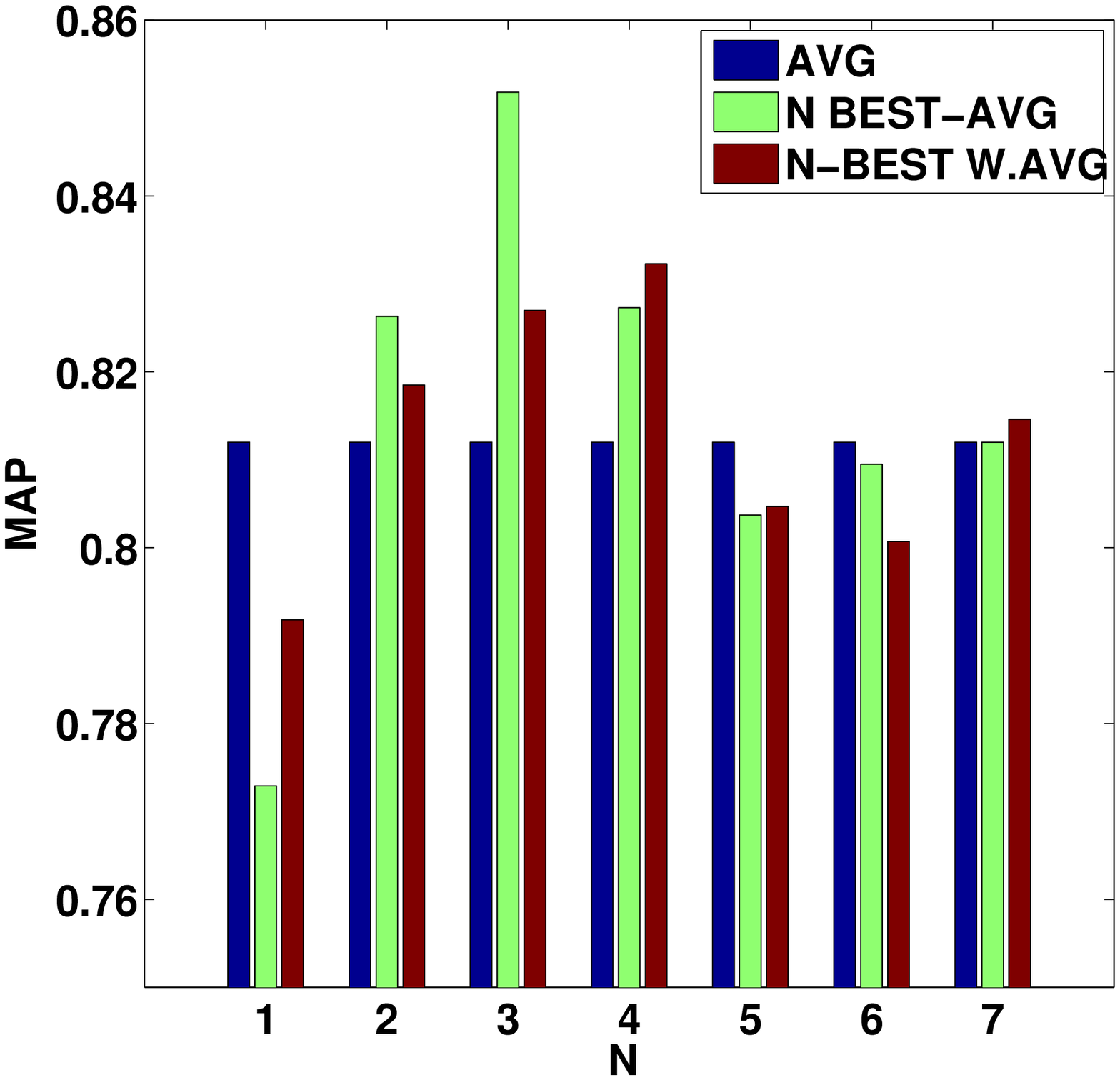}
\caption{MAP results on Set 1 with average fusion of $N$-best scores ($N$-BEST AVG) and weighted fusion of $N$-best scores ($N$-BEST WAVG), as a function of $N$. (a) Left: $3$ classifiers noisy ($c=3$)(b) Right: $4$ classifiers noisy ($c=4$)}
%Horizontal lines show average fusion (avg.), clarity-optimized weighted average  (w.avg.), and optimized raw clarity (RCL).}
\label{fig:nbestnoisy}
\vspace{-0.1in}
\end{figure}

The results for larger number of corrupted classifiers ($c=3$ and $c =4$) are shown as bar plots in Figure \ref{fig:nbestnoisy}. The left panel in Figure \ref{fig:nbestnoisy} shows MAP values for different $N$-best schemes when 3 classifiers are degraded $(c=3)$. It is again clear that the $N$- best selection is robust to noisy classifiers as the performance is sustained to a greater extent. For average fusion the performance drops by $4.03\%$ (from $89.76$ to $85.73$) in terms of MAP. The numbers for $N=3$ for $N$-BEST-AVG and $N$-BEST W.AVG are $87.56\%$ and $89.82\%$ respectively showing that performance is sustained to a much greater extent using the learned weights compared to average fusion. Similar higher MAPs are found for $N=4$ as well. Plots for $c=4$ are shown in the right panel in Figure \ref{fig:nbestnoisy}. In this case the MAP numbers for AVG., $N$-BEST-AVG and $N$-BEST W.AVG with $N=3$ are $81.20\%$, $85.18$, $82.70\%$ respectively. Similar superior performance is observed with $N=4$. This clearly shows that the weight learning algorithm can indeed be used for $N$-best classifier selection which can sustain performance even if some of the classifiers are noisy in the mix.  Plots for other sets are provided in the supplementary material. 

%\emph{This also shows that the algorithm will remain robust in situations where the test set is very large and diverse as compared to the training data on which classifiers were trained. Large and diverse test set can result in erratic behavior of classifier like the present simulated scenario.} 

\section{Conclusions and Discussion}
The results indicate that the proposed fusion method is indeed able to achieve improved results over average fusion, showing its promise. Results also showed that the proposed raw clarity based ranking is a valid metric for ranking instances. In fact it outperformed weighted scoring methods. For several flower classes across different sets, $2-5\%$ absolute improvement in AP is observed using $RCL$ for ranking instances. It is interesting that this score is based primarily on rank order and has no direct probability-based interpretation. Notably though, it is demonstrated that a score that is obtained by unsupervised optimization over the test data is able  to provide improvements over average fusion. This opens up the possibility of an unsupervised weight learning method which can outperform the state of art fusion strategies. The advantages of instance specific unsupervised weight learning are manifold. No held-out set is needed in the optimization process, reducing generalization concerns; also features such as the ability to perform $N$-best selection make the process robust to noise in the outputs of the classifiers.    

The greater benefit from the method is its ability to accurately identify the most promising classifiers to combine, and eliminate noisy classifiers from contention in an {\em instance-specific} manner. This shows that the proposed algorithm can be applied to situations where the test set is large and diverse and it is expected that some classifiers can behave erratically for some test points. We are able to choose the best set of classifiers for {\em each instance} with remarkable consistency. This can, in turn, result in significant improvement in performance, for instance the AP for the class ``Pansy'', an {\em absolute} improvement of 4\%-9\% is achieved in the different sets. In the noise-tolerance study we saw that the $N$-best selection on the noisy classifiers using our proposed method can outperform average fusion by a huge margin in terms of MAP. 

Many avenues remain for investigation. The performance is heavily dependent on optimal choice of $\alpha$ -- while the best $\alpha$ provided by an oracle will result in large improvements in {\em every} case, optimizing $\alpha$ over a held-out development set is unable to find the best $\alpha$ in all cases. For the classifier selection case, while there is considerable latitude in the choice of $N$, the optimal value of $N$ must be identified from a development set. 

From theoretical perspective we need to investigate matters such as the optimal selection of labeled training instances to compute clarity. Another candidate for investigation is the objective function itself: enhancing it with regularizers, {\em e.g.} imposing sparsity on weights. Transductive learning methods that jointly optimize the individual test instances while ensuring that instances with similar scores achieve similar results are likely to result in further improvements. Among work in progress is also a formal proof that the algorithm will always lead to convergence to the best possible clarity given the training set $X$ and score vector $\vec{x_u}$.

{\small
\bibliographystyle{ieee}
\bibliography{references}
}

\end{document}